\documentclass[10pt,twocolumn,letterpaper]{article}

\usepackage[pagenumbers]{cvpr} 

\definecolor{cvprblue}{rgb}{0.21,0.49,0.74}
\usepackage[pagebackref,breaklinks,colorlinks,allcolors=cvprblue]{hyperref}
\usepackage{algorithm}
\usepackage{algpseudocode}
\usepackage{amsmath}

\def\paperID{*****} 
\def\confName{CVPR}
\def\confYear{2025}

\title{sEEG-based Encoding for Sentence Retrieval:\\
A Contrastive Learning Approach to Brain-Language Alignment}

\author{Yijun Liu\\
Ming Hsieh Department of Electrical and Computer Engineering\\University of Southern California \\
{\tt\small yijunl@usc.edu}
}

\begin{document}
\maketitle

\begin{abstract}
Interpreting neural activity through meaningful latent representations remains a complex and evolving challenge at the intersection of neuroscience and artificial intelligence. We investigate the potential of multimodal foundation models to align invasive brain recordings with natural language. We present \textbf{SSENSE}, a contrastive learning framework that projects single-subject stereo-electroencephalography (sEEG) signals into the sentence embedding space of a frozen CLIP model, enabling sentence-level retrieval directly from brain activity. SSENSE trains a neural encoder on spectral representations of sEEG using InfoNCE loss, without fine-tuning the text encoder. We evaluate our method on time-aligned sEEG and spoken transcripts from a naturalistic movie-watching dataset. Despite limited data, SSENSE achieves promising results, demonstrating that general-purpose language representations can serve as effective priors for neural decoding.
\end{abstract}

\footnotetext{This work has been accepted for poster presentation at the CVPR 2025 Workshop on Multimodal Foundation Models (MMFM3).}    
\section{Introduction}
\label{sec:intro}

The ability to decode internal mental content from brain activity—popularly referred to as "mind-reading"—stands as a key challenge at the intersection of cognitive neuroscience and artificial intelligence. Recent breakthroughs in foundation models have enabled powerful cross-modal alignment between modalities such as vision and language, as demonstrated by CLIP \cite{clip} and ALIGN \cite{align}. However, the extension of these capabilities to neural data, particularly high-temporal-resolution signals like stereo-electroencephalography (sEEG), remains underexplored.

In this work, we investigate how to leverage foundation models for brain-to-language alignment. We propose SSENSE (Subject-wise sEEG-based Encoding for Sentence Retrieval), a framework that maps sEEG recordings into the sentence embedding space of a frozen CLIP model. SSENSE uses spectrogram representations of sEEG, processed by a neural encoder trained with a contrastive loss to align neural signals with their corresponding natural language descriptions. The objective is to enable zero-shot sentence retrieval from brain activity by grounding it in a shared semantic space.

We evaluate SSENSE on a dataset comprising time-aligned sEEG recordings and movie transcript sentences collected from a single subject. Despite the limited amount of data, our method achieves promising results on the sentence retrieval task, demonstrating that frozen multimodal language models can serve as effective priors for neural decoding. Our findings provide initial evidence that contrastive brain-language alignment is feasible through embedding into pretrained semantic spaces. This opens new directions for integrating brain signals into the broader landscape of foundation models, with potential applications in cognitive neuroscience, language understanding, and assistive brain-computer interfaces.

\section{Methods}
\label{sec:methods}

\begin{figure*}
  \centering
    \includegraphics[width=\linewidth, height=6.24cm]{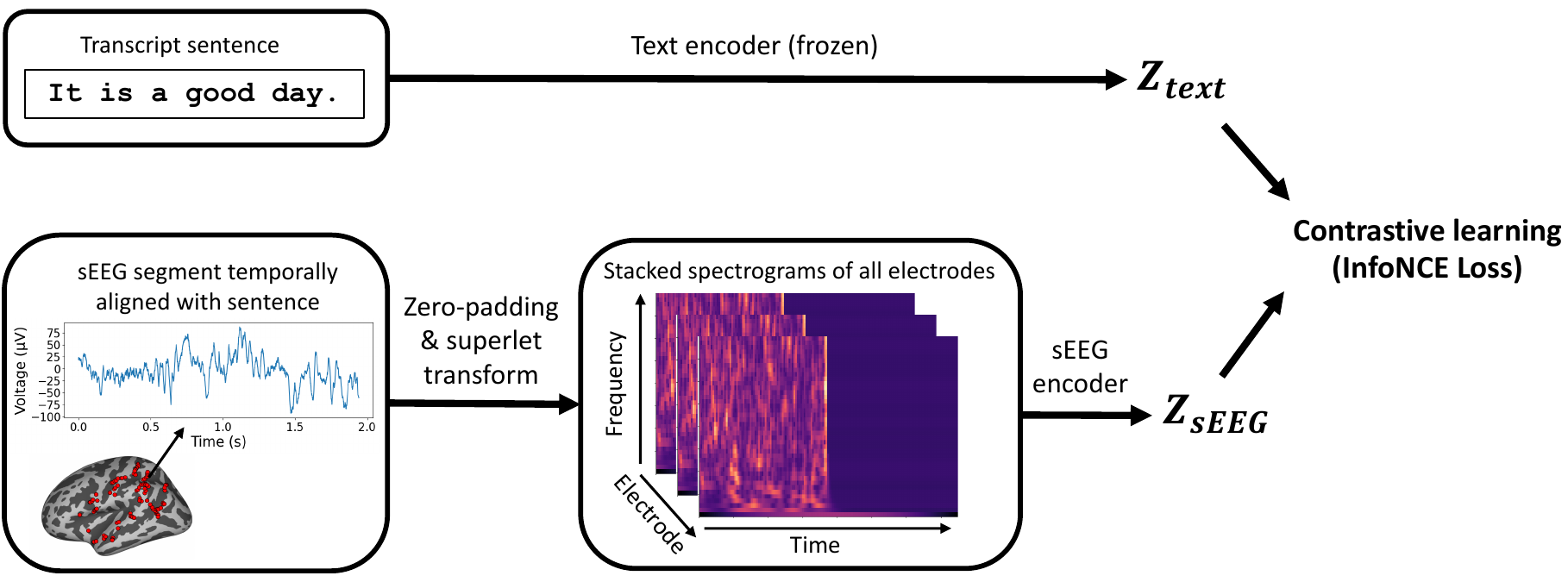}
  \caption{Proposed SSENSE multimodal framework for aligning sEEG signals with natural language. sEEG segments, zero-padded and transformed via the superlet method into time-frequency representations, are encoded using a dedicated sEEG encoder. Sentence embeddings are obtained from the frozen text encoder of CLIP. The model is trained using a contrastive InfoNCE loss to align sEEG and text representations in a shared embedding space.}
  \label{fig:framework}
\end{figure*}

We train a sEEG encoder to align its neural embeddings with the corresponding sentence embeddings that were presented to the subject during movie watching. The overall framework is illustrated in Fig.~\ref{fig:framework}.




\subsection{sEEG Preprocessing}
Raw sEEG signals were first converted into time-frequency representations (spectrograms) using the superlet transform \cite{superlet}, a spectral analysis method previously employed in BrainBERT \cite{brainbert} for self-supervised learning on sEEG data. To standardize input lengths across samples, sEEG segments corresponding to shorter sentences were zero-padded along the time dimension to a fixed length of 8,200 time points (approximately 4 seconds).

\subsection{sEEG Encoder}
To project sEEG recordings into the CLIP sentence embedding space, we design a neural encoder that operates on spectrogram representations of the time-series neural signals. The encoder maps multi-electrode inputs into fixed-dimensional vectors aligned with CLIP’s 512-dimensional sentence embedding space. Given the visual structure of spectrograms, we adopt a modified ResNet-18 architecture \cite{resnet}, pretrained on ImageNet. To accommodate single-channel spectrogram inputs, we adapt the first convolutional layer to accept one input channel, and replace the original classification head with a linear projection that outputs 512-dimensional embeddings. Spectrograms from each electrode are processed independently by the encoder, and the resulting per-electrode embeddings are aggregated via mean pooling across electrodes to form a unified representation per sample. This design enables the model to capture both spatial and temporal patterns within each electrode’s signal for contrastive alignment with natural language.

\subsection{Text Encoder}
We leverage the pretrained CLIP \cite{clip} text encoder (ViT-B/32) as a frozen module to extract semantic representations of natural language stimuli. Each sentence is tokenized using the CLIP tokenizer and passed through the encoder to produce a 512-dimensional embedding. To ensure stable training and effective alignment in the shared embedding space, we apply L2 normalization to all text embeddings. Importantly, the CLIP model remains fixed throughout training, serving as a consistent target for contrastive alignment with brain-derived neural representations. This strategy enables the model to exploit rich language priors captured by large-scale vision-language pretraining while focusing learning capacity on the neural encoder.

\subsection{Data Augmentation Strategies}
To enhance generalization, we explore data augmentation strategies inspired by the time-frequency masking techniques introduced in BrainBERT~\cite{brainbert}. Specifically, we apply time-frequency masking directly to the spectrogram inputs and randomly mask a subset of electrode channels during training. These augmentations aim to improve the model’s robustness to noise and inter-trial variability inherent in neural recordings. Pseudocode for both augmentation strategies is provided below.

\label{sec:data_augmentation}

\begin{algorithm}[H]
\caption{Time-Frequency Masking}
\begin{algorithmic}[1]
\Require Input tensor $x \in \mathbb{R}^{B \times E \times C \times F \times T}$, frequency mask ratio $r_f$, time mask ratio $r_t$
\For{$b = 1$ to $B$}
    \If{$r_f > 0$}
        \State Sample $M_f = \lfloor r_f \cdot F \rfloor$ random frequency indices
        \State Set $x[b, :, :, M_f, :] = 0$
    \EndIf
    \If{$r_t > 0$}
        \State Sample $M_t = \lfloor r_t \cdot T \rfloor$ random time indices
        \State Set $x[b, :, :, :, M_t] = 0$
    \EndIf
\EndFor
\Ensure Masked tensor $x$
\end{algorithmic}
\end{algorithm}

\begin{algorithm}[H]
\caption{Electrode Masking}
\begin{algorithmic}[1]
\Require Input tensor $x \in \mathbb{R}^{B \times E \times C \times F \times T}$, mask ratio $r$
\For{$b = 1$ to $B$}
    \State Sample $M = \lfloor r \cdot E \rfloor$ random electrode indices
    \State Set $x[b, \text{masked electrodes}, :, :, :] = 0$
\EndFor
\Ensure Masked tensor $x$
\end{algorithmic}
\end{algorithm}

\begin{table*}[t]
  \centering
  \begin{tabular}{@{}lcccc@{}}
    \toprule
    Method & Recall@1 & Recall@10 & Recall@50 & MRR\\
    \midrule
    SSENSE &  &  &  &  \\
    \quad no masking & 1.2 $\pm$ 0.54 & 10.68 $\pm$ 1.40 & 40.58 $\pm$ 2.29 & 0.0498 $\pm$ 0.0049 \\
    \quad with electrode masking & 0.72 $\pm$ 0.25 & 9.00 $\pm$ 1.21 & 40.00 $\pm$ 1.76 & 0.0446 $\pm$ 0.0044 \\
    \quad with frequency and time masking & 0.99 $\pm$ 0.44 & 9.65 $\pm$ 1.41 & 38.69 $\pm$ 2.75 & 0.0475 $\pm$ 0.0062 \\
    \quad with electrode and frequency and time masking & 1.06 $\pm$ 0.57 & 9.38 $\pm$ 1.78 & 40.37 $\pm$ 1.78 & 0.0465 $\pm$ 0.0077 \\
    Random sentence retrieval & 0.34 & 3.42 & 17.12 & 0.0214 \\
    \bottomrule
  \end{tabular}
  \caption{Sentence retrieval performance from sEEG signals. Mean and standard deviation were computed over 10 random seeds with different dataset splits. The training, validation, and test sets were fixed at 60\%, 20\%, and 20\%, respectively.}
  \label{tab:eval}
\end{table*}

\subsection{Loss function}

We used InfoNCE \cite{infonce} to maximize the consine similarity between the positive sEEG-sentence pairs (sEEG and sentence recorded at the same time). 

Given a batch of $N$ paired samples $\{(z_i^{\text{sEEG}}, z_i^{\text{text}})\}_{i=1}^N$, where $z_i^{\text{sEEG}}$ is the embedding of the sEEG input and $z_i^{\text{text}}$ is the embedding of the corresponding text (from a frozen encoder), the InfoNCE loss for a positive pair $(i)$ is:

\begin{equation}
\mathcal{L}_{i} = -\log \frac{\exp\left(\frac{\text{sim}(z_i^{\text{sEEG}}, z_i^{\text{text}})}{\tau}\right)}{
\sum_{j=1}^N \exp\left(\frac{\text{sim}(z_i^{\text{sEEG}}, z_j^{\text{text}})}{\tau}\right)}
  \label{eq:single_loss}
\end{equation}

where $\text{sim}(a, b) = \frac{a^\top b}{\|a\|\|b\|}$ is the cosine similarity, and $\tau$ is the temperature hyperparameter, fixed at 0.07. The final loss is averaged over the batch:

\begin{equation}
  \mathcal{L}_{\text{InfoNCE}} = \frac{1}{N} \sum_{i=1}^N \mathcal{L}_i
  \label{eq:loss}
\end{equation}

Optimization was performed using the Adam optimizer with a learning rate of 0.0005. Early stopping was applied after 5 consecutive epochs without improvement in validation Recall@10.

\section{Experiments}
\label{sec:exp}
\subsection{Dataset}
We conducted experiments using sEEG recordings from a single subject (female, 16 years old) provided by the Brain Treebank dataset \cite{braintreebank}. The subject watched the audiovisual movie Ant-Man, for which aligned transcripts and trigger-based timestamps were available, allowing precise synchronization between the sEEG signals and the movie’s linguistic content. The full movie spans approximately 1.8 hours and contains 1,558 transcribed sentences.

To reduce computational complexity and ensure consistency, we filtered out overly long sentences, retaining only those with durations under 4 seconds. Additionally, each sEEG segment was extended to include 0.5 seconds of context before the sentence onset and 1 second after the sentence offset.

After preprocessing, we obtained 1,454 valid sEEG–sentence pairs. This dataset was divided into 60\% for training (872 samples), 20\% for validation (290 samples), and 20\% for testing (292 samples). To account for potential variability due to data splits, we repeated the experiments across 10 different random seeds, each representing a distinct partitioning of the dataset. In all cases, the train/validation/test proportions were held constant. Importantly, test samples were kept completely unseen during the contrastive training stage to ensure unbiased evaluation.

We experimented with various masking percentages for both time-frequency and electrode masking. The final masking configurations used for testing were selected based on the best Recall@10 performance on the validation set.

\subsection{Zero-shot sEEG to transcript sentence retrieval}

The pretraining objective is to align sEEG signal embeddings with CLIP's text embedding space using a contrastive learning framework. By jointly training the sEEG encoder with sentence embeddings from a frozen CLIP model, the system learns to project neural activity—recorded during naturalistic movie watching—into a shared semantic space with language. This alignment enables zero-shot retrieval: neural representations observed at test time can be directly compared to candidate sentence embeddings without any additional fine-tuning.

To assess the effectiveness of this alignment, we conduct a retrieval task in which, given an sEEG segment recorded while the subject listened to a sentence, the model must identify the corresponding sentence from a set of candidates. Retrieval is performed by computing cosine similarity between the sEEG embedding and each candidate sentence embedding in the shared space, ranking them accordingly. This task reflects a practical neural decoding scenario—recovering linguistic content directly from brain activity without explicit supervision or retraining.

Performance is measured using standard retrieval metrics, including Recall@$k$ and Mean Reciprocal Rank (MRR), which quantify how highly the correct sentence is ranked among retrieved candidates. Crucially, because the CLIP text encoder remains frozen and the sentence candidates are unseen during training, this evaluation serves as a rigorous test of the generalizability and semantic grounding of the learned neural representations.

In addition to reporting the mean and standard deviation across 10 trials with different random seeds, we perform statistical analysis to assess significance. Specifically, one-sample \textit{t}-tests are used to compare each SSENSE variant against the random retrieval baseline, and paired-sample \textit{t}-tests are used to compare performance between different SSENSE configurations across trials.

\section{Results}

Table~\ref{tab:eval} presents the sentence retrieval performance of SSENSE under four masking configurations: no masking, electrode masking, frequency-time masking, and combined electrode and frequency-time masking. All variants significantly outperform the random retrieval baseline, demonstrating that SSENSE effectively captures meaningful correspondences between sEEG signals and language.

Among the variants, the model without masking achieves the highest overall performance, with 1.2\% Recall@1, 10.68\% Recall@10, and an MRR of 0.0498. Compared to frequency-time masking, the performance differences are not statistically significant across metrics (all $p > 0.08$), suggesting that frequency-time masking does not improve or degrade performance. In contrast, electrode masking shows a statistically significant drop in performance relative to no masking in Recall@1 ($p = 0.045$), Recall@10 ($p = 0.011$), and MRR ($p = 0.028$), indicating that masking entire channels can obscure important spatial information. The fully masked variant (electrode + frequency-time) does not offer any performance advantage and performs comparably to the frequency-time only variant (all $p > 0.12$).

These results suggest that while frequency-time masking can be used without significantly affecting retrieval quality, electrode masking---especially when combined with other augmentations---may hinder model performance by removing essential neural information. Careful design of data augmentation strategies is thus critical in preserving relevant signal features for fine-grained multimodal alignment.

\section{Conclusion and Discussion}

We presented SSENSE, a contrastive framework aligning intracranial sEEG recordings with natural language in a shared semantic space. Using a frozen CLIP text encoder and InfoNCE-based sEEG encoder, SSENSE enables zero-shot sentence retrieval from brain activity without task-specific tuning. Results show consistent outperformance over random baselines, highlighting the feasibility of decoding language from neural signals in naturalistic settings. This work is an early step toward brain-grounded foundation models. Future directions include scaling to multi-subject data, integrating visual inputs, and leveraging larger language models. Code will be released after follow-up work is completed.

{
    \small
    \bibliographystyle{ieeenat_fullname}
    \bibliography{main}
}


\end{document}